# From Constraints to Resolution Rules
# Part II : chains, braids, confluence and T&E

Denis Berthier
Institut Telecom ; Telecom & Management SudParis
9 rue Charles Fourier, 91011 Evry Cedex, France

**Abstract: In this Part II, we apply the general theory developed in Part I to a detailed analysis of the Constraint Satisfaction Problem (CSP). We show how specific types of resolution rules can be defined. In particular, we introduce the general notions of a chain and a braid. As in Part I, these notions are illustrated in detail with the Sudoku example - a problem known to be NP-complete and which is therefore typical of a broad class of hard problems. For Sudoku, we also show how far one can go in "approximating" a CSP with a resolution theory and we give an empirical statistical analysis of how the various puzzles, corresponding to different sets of entries, can be classified along a natural scale of complexity. For any CSP, we also prove the confluence property of some Resolution Theories based on braids and we show how it can be used to define different resolution strategies. Finally, we prove that, in any CSP, braids have the same solving capacity as Trial-and-Error (T&E) with no guessing and we comment this result in the Sudoku case.**
**Keywords:** constraint satisfaction problem, knowledge engineering, modelling and simulation, production system, resolution rule, chains, braids, confluence, Trial-and Error, Sudoku solving, Sudoku rating.

## I. INTRODUCTION

In Part I of this paper, which is an inescapable pre-requisite to the present Part II, the Constraint Satisfaction Problem (CSP) [1, 2] was analysed in a new general framework based on the idea of a constructive, pattern-based solution and on the concepts of a candidate and a resolution rule. Here we introduce several additional notions valid for any CSP, such as those of a chain, a whip and a braid. We show how these patterns can be the basis for new general and powerful kinds of resolution rules. All of the concepts defined here are straightforward generalisations (and formalisations) of those we introduced in the Sudoku case [3, 4]. Because of space constraints, we formulate our concepts only in plain English but they can easily be formalised with logical formulæ using the basic concepts introduced in Part I.

We give a detailed account of how these general notions can be applied to Sudoku solving. Sudoku is a very interesting problem for several reasons: 1) it is known to be NP-complete [5] (more precisely, the CSP family Sudoku(n) on square grids of size n for all n is NP-complete); 2) nevertheless, it is much easier to study than Chess or Go; 3) a Sudoku grid is a particular case of Latin Squares; Latin Squares are more elegant, from a mathematical point of view, because there is a complete symmetry between all the variables: rows, columns, numbers; in Sudoku, the constraint on blocks introduces some apparently mild complexity which makes it more exciting for players; 4) there are millions of Sudoku players all around the world and many forums, with a lot of cumulated experience available – including generators of random puzzles. For all these reasons, we chose the Sudoku example instead of the more "mathematically correct" Latin Squares CSP.

Whereas sections II and III define the general chains and the elementary bivalue chains, sections IV and V introduce three powerful generalisations of bivalue chains: zt-chains, zt-whips and zt-braids. Section VI defines the very important property of confluence and the notion of a resolution strategy; it proves the confluence property of natural braid resolution theories. Finally, section VII proves that braids have the same solving potential as Trial-and-Error with no guessing.

## II. CHAINS IN A GENERAL CSP

Definition: two different candidates of a CSP are *linked* by a direct contradiction (or simply linked) if some of the constraints of the CSP directly prevents them from being true at the same time *in any knowledge state* in which they are present (the fact that this notion does not depend on the knowledge state is fundamental for the sequel). For any CSP, two different candidates for the same variable are always linked; but there are generally additional direct contradictions; as expliciting them is part of modelling the CSP, we consider them as givens of the CSP and we introduce a basic predicate "linked$_{ij}$($x_i$, $x_j$)" to express them, for each couple of CSP variables $X_i$ and $X_j$. In Sudoku, two different candidates $n_1r_1c_1$ and $n_2r_2c_2$ are linked and we write linked($n_1r_1c_1$, $n_2r_2c_2$), if:
($n_1 \neq n_2$ & $r_1c_1 = r_2c_2$) or ($n_1 = n_2$ & share-a-unit($r_1c_1$, $r_2c_2$)).

Definition: an *Elementary Constraint Propagation* rule is a resolution rule expressing such a direct contradiction. For any CSP, we note ECP the set of all its elementary constraints propagation rules. An ECP rule has the general form:
$\forall x_i \forall x_j$ value$_i(x_i)$ & linked$_{ij}(x_i, x_j)$ => ¬ cand$_j(x_j)$.

Chains (together with whips and braids) appear to be the main tool for dealing with hard instances of a CSP.

Definitions: *a chain of length n* is a sequence $L_1, R_1, L_2, R_2, \ldots L_n, R_n$, of $2n$ *different* candidates for possibly different variables such that: for any $1 \leq k \leq n$, $R_k$ is linked to $L_k$ and for any $1 \leq k \leq n$, $L_k$ is linked to $R_{k-1}$. A *target of a chain* is any candidate that is linked to both its first and its last candidates.

Of course, these conditions are not enough to ensure the existence of an associated resolution rule concluding that the target can be eliminated. Our goal is now to define more specific types of chains allowing such a conclusion.

### III. BIVALUE-CHAINS IN A GENERAL CSP

*A. Bivalue-chains in a general CSP*

Definition: a variable is called *bivalue* in a knowledge state KS if it has exactly two candidates in KS.

Definition and notation: in any CSP, *a bivalue-chain of length n* is a chain of length n: $L_1, R_1, L_2, R_2, \ldots L_n, R_n$, such that, additionally: for any $1 \leq k \leq n$, $L_k$ and $R_k$ are candidates for the same variable, and this variable is bivalue. A bivalue-chain is written symbolicaly as: $\{L_1\ R_1\}$ - $\{L_2\ R_2\}$ - $\ldots$ - $\{L_n\ R_n\}$, where the curly braces recall that the two candidates are relative to the same variable.

***bivalue-chain rule for a general CSP: in any knowledge state of any CSP, if Z is a target of a bivalue-chain, then it can be eliminated (formally, this rule concludes ¬Z).***

Proof: the proof is short and obvious but it will be the basis for all our forthcoming chain and braid rules.

If Z was true, then $L_1$ would be false; therefore $R_1$ would have to be the true value of the first variable; but then $L_2$ would be an impossible value for the second variable and $R_2$ would be its true value….; finally $R_n$ would be true in the last cell; which contradicts Z being true. Therefore Z can only be false. qed.

*B. xy-chains in Sudoku*

We shall adopt the following definitions [1]. Two different rc-cells are linked if they share a unit (i.e. they are in the same row, column or block). A *bivalue cell* is an rc-cell in which there are exactly two candidates (here considered as numbers in these cells). An *xy-chain of length n* is a sequence of n different bivalue rc-cells (each represented by a set notation: $\{\ldots\}$) such that each (but the first) is linked to the previous one (represented by a "-") , with contents: $\{a_1\ a_2\}$ - $\{a_2\ a_3\}$ - $\ldots$ $\{a_n\ a_1\}$. A *target* of the above xy-chain is a number $a_1$ in a cell that is linked to the first and last ones. xy-chains are the most classical and basic type of chains in Sudoku. Our presentation is non standard, but equivalent to the usual ones [6, 7].

***Classical xy-chain rule in Sudoku: if Z is a target of an xy-chain, then it can be eliminated.***

*C. nrc-chains in Sudoku*

The above definition of an xy-chain in Sudoku is the traditional one and it corresponds to the general notion of a bivalue-chain in any CSP, when we consider only the natural variables $X_{rc}$ and $X_{bs}$ of the Sudoku CSP. But it is not as general as it could be. To get the most general definition, we must consider not only the "natural" $X_{rc}$ variables but also the corresponding $X_{rn}$, $X_{cn}$ and $X_{bn}$ variables, as introduced in Part I, with $X_{rc} = n \Leftrightarrow X_{rn} = c \Leftrightarrow X_{cn} = r \Leftrightarrow X_{bn} = s$, whenever correspondence(r, c, b, s) is true. The notion of bivalue is meaningful for each of these variables. And, when we use all these variables instead of only the $X_{rc}$, we get a more general concept of bivalue-chains, which we called nrc-chains in [4] and which are a different view of some classical Nice Loops [6, 7]. The notion of "bivalue" for these non-standard variables corresponds to the classical notion of conjugacy in Sudoku – but, from the point of view of the general theory, there is no reason to make any difference between "bivalue" and "conjugate". In the sequel, we suppose that we use all the above variables.

***Classical nrc-chain rule in Sudoku: any target of an nrc-chain can be eliminated.***

### IV. THE Z- AND T- EXTENSIONS OF BIVALUE-CHAINS IN A CSP

We first introduced the following generalisations of bivalue-chains in [3], in the Sudoku context. But everything works similarly for any CSP. It is convenient to say that a candidate C is *compatible* with a set S of candidates if it is not linked to any element of S.

*A. t-chains, z-whips and zt-whips in a general CSP*

The definition of a bivalue-chain can be extended in different ways, as follows.

Definition: a *t-chain* of length n is a chain $L_1, R_1, L_2, R_2, \ldots L_n, R_n$, such that, additionally, for each $1 \leq k \leq n$:

- $L_k$ and $R_k$ are candidates for the same variable,
- $R_k$ is the only candidate for this variable compatible with the previous right-linking candidates.

***t-chain rule for a general CSP: in any knowledge state of any CSP, any target of a t-chain can be eliminated (formally, this rule concludes ¬Z).***

For the z- extension, it is natural to introduce *whips* instead of chains. Whips are also more general, because they are able to catch more contradictions than chains. A *target of a whip* is required to be linked to its first candidate, not necessarily to its last.

Definition: given a candidate Z (which will be the target), a *z-whip* of length n built on Z is a chain $L_1, R_1, L_2, R_2, \ldots, L_n$ (notice that there is no $R_n$), such that, additionally:

– for each $1 \leq k < n$, $L_k$ and $R_k$ are candidates for the same variable,

– $R_k$ is the only candidate for this variable compatible with Z (apart possibly for $L_k$),

– for the same variable as $L_n$, there is no candidate compatible with the target.

Definition: given a candidate Z (which will be the target), a *zt-whip* of length n built on Z is a chain $L_1$, $R_1$, $L_2$, $R_2$, …. $L_n$ (notice that there is no $R_n$), such that, additionally:

– for each $1 \leq k < n$, $L_k$ and $R_k$ are candidates for the same variable,

– $R_k$ is the only candidate for this variable compatible with Z and the previous right-linking candidates,

– for the same variable as $L_n$, there is no candidate compatible with the target and the previous right-linking candidates.

***z- and zt-whip rules for a general CSP: in any knowledge state of any CSP, if Z is a target of a z- or a zt- whip, then it can be eliminated (formally, this rule concludes ¬Z).***

Proof: the proof can be copied from that for the bivalue-chains. Only the end is slightly different. When variable $L_n$ is reached, it has negative valence. With the last condition on the whip, it entails that, if the target was true, there would be no possible value for the last variable.

Remark: although these new chains or whips seem to be straightforward generalisations of bivalue-chains, their solving potential is much higher. Soon, we'll illustrate this with the Sudoku example.

Definition: in any of the above chains or whips, a value of the variable corresponding to candidate $L_k$ is called a t- (resp. z-) candidate if it is incompatible with the previous right-linking (i.e. the $R_i$) candidates (resp. with the target).

*B. zt-whip resolution theories in a general CSP*

We are now in a position to define an increasing sequence of resolution theories based on zt-whips: BRT is the Basic Resolution Theory defined in Part I. $L_1$ is the union of BRT and the rule for zt-whips of length 1. For any n, $L_{n+1}$ is the union of $L_n$ with the rule for zt-whips of length n+1. $L_∞$ is also defined, as the union of all the $L_n$. In practice, as we have a finite number of variables in finite domains, $L_∞$ will be equal to some $L_n$.

*C. t-whips, z-whips and zt-whips in Sudoku*

In Sudoku, depending on whether we consider only the "natural" $X_{rc}$ and $X_{bs}$ variables or also the corresponding $X_{rn}$, $X_{cn}$ and $X_{bn}$ variables, we get xyt-, xyz- and xyzt- whips or nrct-, nrcz- and nrczt- whips. In the Sudoku case, we have programmed all the above defined rules for whips in our SudoRules solver, a knowledge based system, running indifferently on the CLIPS [8] or the JESS [9] inference engine.

This allowed us to obtain the following statiscal results.

*D. Statistical results for the Sudoku nrczt-whips*

Definition: a puzzle is *minimal* if it has one and only one solution and it would have several solutions if any of its entries was deleted. In statistical analyses, only samples of minimal puzzles are meaningful because adding extra entries would multiply the number of easy puzzles. In general, puzzles proposed to players are minimal.

One advantage of taking Sudoku as our standard example (instead of e.g. Latin Squares) is that there are generators of random minimal puzzles. Before giving our results, it is necessary to mention that there are puzzles of extremely different complexities. With respect to several natural measures of complexity one can use (number of partial chains met in the solution, computation time, …), provided that they are based on resolution rules (instead of e.g. blind search with backtracking), different puzzles will be rated in a range of several orders of magnitude (beyond 13 orders in Sudoku).

The following statistics are relative to a sample of 10,000 puzzles obtained with the suexg [10] random generator. Row 3 of Table 1 gives the total number of puzzles solved when whips of length $\leq$ n (corresponding to resolution theory $L_n$) are allowed; row 2 gives the difference between $L_n$ and $L_{n-1}$. (Of course, in any $L_n$, the rules of BSRT, consisting of ECP, NS, HS and CD are allowed in addition to whips).

| BSRT | L1 | L2 | L3 | L4 | L5 | L6 | L7 |
|---|---|---|---|---|---|---|---|
| 4247 | 1135 | 1408 | 1659 | 1241 | 239 | 56 | 10 |
| 4247 | 5382 | 6790 | 8449 | 9690 | 9929 | 9985 | 9995 |

Table 1: Number of puzzles solved with nrczt-whips of length ≤ n. The 5 remaining puzzles can also be solved with whips, although longer ones.

As these results are obtained from a very large random sample, they show that almost all the minimal puzzles can be solved with nrczt-whips. But they don't allow to conclude for all the puzzles. Indeed, extremely rare cases are known which are not solvable with nrczt-whips only. They are currently the puzzles of interest for researchers in Sudoku solving. But, for the Sudoku player, they are very likely to be beyond his reach, unless radically new types of rules are devised.

V. ZT-BRAIDS IN A GENERAL CSP

We now introduce a further generalisation of whips: braids. Whereas whips have a linear structure (a chain structure), braids have a (restricted) net structure. In any CSP, braids are interesting for three reasons: 1) they have a greater solving potential than whips (at the cost of a more complex structure); 2) resolution theories based on them can be proven to have the very important confluence property, allowing to introduce various resolution strategies based on them; and 3) their scope can be defined very precisely; they can eliminate any candidate that can be eliminated by pure Trial-and-Error (T&E); they can therefore solve any puzzle that can be solved by T&E.

*A. Definition of zt-braids*

Definition: given a target Z, a *zt-braid* of length n built on Z is a sequence of candidates $L_1, R_1, L_2, R_2, \ldots L_n$ (notice that there is no $R_n$), such that:

– for each $1 \leq k \leq n$, $L_k$ is linked either to a previous right-linking candidate (some $R_l$, $l < k$) or to the target (this is the main structural difference with whips),

– for each $1 \leq k < n$, $L_k$ and $R_k$ are candidates for the same variable (they are therefore linked),

– $R_k$ is the only candidate for this variable compatible with the target and the previous right-linking candidates,

– for the variable corresponding to candidate $L_n$, there is no candidate compatible with the target and the previous right-linking candidates.

In order to show the kind of restriction this definition entails, the first of the following two structures can be part of a braid starting with $\{L_1\ R_1\}$ - $\{L_2\ R_2\}$ -…, whereas the second can't:

$\{L_1\ R_1\}$ - $\{L_2\ R_2\ A_2\}$ - … where $A_2$ is linked to $R_1$;

$\{L_1\ R_1\ A_1\}$ - $\{L_2\ R_2\ A_2\}$ - … where $A_1$ is linked to $R_2$ and $A_2$ is linked to $R_1$ but none of them is linked to Z. The only thing that could be concluded from this pattern if Z was true is ($R_1$ & $R_2$) or ($A_1$ & $A_2$), whereas a braid should allow to conclude $R_1$ & $R_2$.

The proof of the following theorem is exactly the same as for whips, thanks to the linear order of the candidates.

***zt-braid rule for a general CSP: in any knowledge state of any CSP, if Z is a target of a zt-braid, then it can be eliminated (formally, this rule concludes ¬Z).***

Braids are a true generalisation of whips. Even in the Sudoku case (for which whips solve almost any puzzle), examples can be given of puzzles that can be solved with braids but not with whips. This will be a consequence of our T&E vs braid theorem.

VI. CONFLUENCE PROPERTY, BRAIDS, RESOLUTION STRATEGIES

*A. The confluence property*

Given a resolution theory T, consider all the strategies that can be built on it, e.g. by defining various priorities on the rules in T. Given an instance P of the CSP and starting from the corresponding knowledge state $KS_P$, the resolution process associated with a stategy S built on T consists of repeatedly applying resolution rules from T according to the additional conditions (e.g. the priorities) introduced by S. Considering that, at any point in the resolution process, different rules from T may be applicable (and different rules will be applied) depending on the chosen stategy S, we may obtain different resolution paths starting from $KS_P$ when we vary S.

Let us define the *confluence property* as follows: a Resolution Theory T for a CSP has the confluence property if, for any instance P of the CSP, any two resolution paths can be extended to meet in a common knowledge sate. In this case, all the resolution paths starting from $KS_P$ and associated with all the stategies built on T will lead to the same final state in ***$KS_P$*** (all explicitly inconsistent states are considered as identical; they mean contradictory constraints). If a resolution theory T doesn't have the confluence property, one must be careful about the order in which he applies the rules. But if T has this property, one may choose any resolution strategy, which makes finding a solution much easier.

*B. The confluence property of zt-braid resolution theories*

As for whips, one can define an increasing sequence of resolution theories based on zt-braids: $M_1$ is the union of BRT and the rule for zt-braids of length 1. (Notice that $M_1 = L_1$). For any n, $M_{n+1}$ is the union of $M_n$ with the rule for zt-braids of length n+1. $M_∞$ is defined as the union of all the $M_n$.

***Theorem: any of the above zt-braid theories has the confluence property.***

Before proving this theorem, we must give a precison about candidates. When one is asserted, its status changes: it becomes a value and it is deleted as a candidate. (The theorem doesn't depend on this but the proof should have to be slightly modified with other conventions).

Let n be fixed. What our proof will show is the following much stronger stability property: for any knowledge state KS, any elimination of a candidate Z that might have been done in KS by a zt-braid B of length n and target Z will always be possible in any further knowledge state (in which Z is still a candidate) using rules from $M_n$ (i.e. for zt-braids of length n or less, together with BRT). For this, we must consider all that can happen to B. Let B be:
$\{L_1\ R_1\}$ - $\{L_2\ R_2\}$ - …. - $\{L_p\ R_p\}$ - $\{L_{p+1}\ R_{p+1}\}$ - … - $L_n$.

If the target Z is eliminated, then our job is done. If Z is asserted, then the instance of the CSP is contradictory. This contradiction will be detected by CD after a series of ECP and S following the braid structure.

If a right-linking candidate, say $R_p$, is eliminated, the corresponding variable has no possible value and we get the shorter braid with target Z: $\{L_1\ R_1\}$ - $\{L_2\ R_2\}$ - …. - $L_p$. If a left-linking candidate, say $L_{p+1}$, is asserted, then $R_p$ can be eliminated by ECP, and we are in the previous case.

If a right-linking candidate, say $R_p$, is asserted, it can no longer be used as an element of a braid. Notice that $L_{p+1}$ and all the t-candidates in cells of B after p that were incompatible with $R_p$, i.e. linked to it, can be eliminated by ECP. Let q be the smallest number greater than p such that, after all these eliminations, cell number q still has a t- or a z- candidate $C_q$; notice that the right-linking candidates in all the cells between p and q-1 can be asserted by S, all the t-candidates in cells after q that were incompatible with either of them can be eliminated by ECP and all the left-linking candidates in all the cells between p and q can be eliminated by ECP. Let k be the largest number k $\leq$ p such that $C_q$ is incompatible with $R_k$ (or q = 0 if C is incompatible only with Z). Then the shorter braid obtained from B by excising cells p+1 to q and by replacing $L_q$ by $C_q$ still has Z has its target and can be used to eliminate it.

Suppose now a left-linking candidate, say $L_p$, is eliminated. Either $\{L_p\ R_p\}$ was bivalue, in which case $R_p$ can be asserted by S and we are in the previous case. Or there remains some t- or z-candidate C for this variable and we can consider the braid, with target Z, obtained by replacing $L_p$ by C. Notice that, even if $L_p$ was linked to $R_{p-1}$, this may not be the case for C; therefore trying to prove a similar theorem for whips would fail here.

If any t- or z- candidate is eliminated, then the basic structure of B is unchanged. If any t- or z- candidate is asserted as a value, then the right-linking candidate of its cell can be eliminated by ECP and we are in one of the previous cases.

As all the cases have been considered, the proof can be iterated in case several of these events have happened to B. Notice that this proof works only because the notion of being linked doesn't depend on the knowledge state.

*C. Resolution strategies*

There are the Resolution Theories defined above and there are the many ways one can use them in practice to solve real instances of a CSP. From a strict logical standpoint, all the rules in a Resolution Theory are on an equal footing, which leaves no possibility of ordering them. But, when it comes to the practical exploitation of resolution theories and in particular to their implementation, e.g. in an inference engine as in our SudoRules solver, one question remains unanswered: can superimposing some ordering on the set of rules (using priorities or "saliences") prevent us from reaching a solution that the choice of another ordering might have made accessible? With resolution theories that have the confluence property such problems cannot appear and one can take advantage of this to define different resolution strategies.

*Resolution strategies* based on a resolution theory T can be defined in different ways and may correspond to different goals:
  – implementation efficiency;
  – giving a preference to some patterns over other ones: preference for chains over zt-whips and/or for whips over braids;
  – allowing the use of heuristics, such as focusing the search on the elimination of some candidates (e.g. because they correspond to a bivalue variable or because they seem to be the key for further eliminations); but good heursitics are hard to define.

VII. BRAIDS VS TRIAL-AND-ERROR IN A GENERAL CSP

*A. Definition of the Trial and Error procedure (T&E)*

Definition: given a resolution theory T, a knowledge state KS and a candidate Z, *Trial and Error based on T for Z, T&E(T, Z)*, is the following procedure (notice: a procedure, not a resolution rule): make a copy KS' of KS; in KS', delete Z as a candidate and assert it as a value; in KS', apply repeatedly all the rules in T until quiescence; if a contradiction is obtained in KS', then delete Z from KS; otherwise, do nothing.

Given a fixed resolution theory T and any instance P of a CSP, one can try to solve it using only T&E(T). We say that P can be solved by T&E(T) if, using the rules in T any time they can be applied plus the procedure T&E(T, Z) for some remaining candidate Z every time no rule from T can be applied, a solution of P can be obtained. When T is the BRT of our CSP, we simply write T&E instead of T&E(T).

As using T&E leads to examining arbitrary hypotheses, it is often considered as blind search. But notice nevertheless that it includes no "guessing": if a solution is obtained in an auxiliary state KS', then it is not taken into account, as it would in standard structured search algorithms.

*B. zt-braids versus T&E theorem*

It is obvious that any elimination that can be made by a zt-braid can be made by T&E. The converse is more interesting.

**Theorem: for any instance of any CSP, any elimination that can be made by T&E can be made by a zt-braid. Any instance of a CSP that can be solved by T&E can be solved by zt-braids.**

Proof: Let Z be a candidate eliminated by T&E using some auxiliary knowledge state KS'. Following the steps of T&E in KS', we progressively build a zt-braid in KS with target Z. First, remember that BRT contains three types of rules: ECP (which eliminates candidates), $S_k$ (which asserts a value for the k-th variable of the CSP) and $CD_k$ (which detects a contradiction on variable $X_k$). Consider the first step of T&E which is the application of some $S_k$ in KS', thus asserting some $R_1$. As $R_1$ was not in KS, there must have been some elimination of a candidate, say $L_1$, made possible in KS' by the assertion of Z, which in turn made the assertion of $R_1$ possible in KS'. But if $L_1$ has been eliminated in KS', it can only be by ECP and because it is linked to Z. Then $\{L_1\ R_1\}$ is the first cell of our zt-braid in KS. (Notice that there may be other z-candidates in cell $\{L_1\ R_1\}$, but this is pointless, we can choose any of them as $L_1$ and consider the remaining ones as z-candidates). The sequel is done by recursion. Suppose we have built a zt-braid in KS corresponding to the part of the T&E procedure in KS' until its n-th assertion step. Let $R_{n+1}$ be the next candidate asserted in KS'. As $R_{n+1}$ was not asserted in KS, there must have been some elimination in KS' of a candidate, say $L_{n+1}$, made possible by the assertion in KS' of Z or of some of the previous $R_k$, which in turn made the assertion of $R_{n+1}$ possible in KS'. But if $L_{n+1}$ has been eliminated in KS', it can only be by ECP and because it is linked to Z or to some of the previous $R_k$, say C. Then our partial braid in KS can be extended with cell {$L_{n+1}$ $R_{n+1}$}, with $L_{n+1}$ linked to C.

End of the procedure: either no contradiction is obtained by T&E and we don't have to care about any braid in KS, or a contradiction is obtained. As only ECP can eliminate a candidate, a contradiction is obtained when the last asserted value, say $R_{n-1}$, eliminates (via ECP) a candidate, say $L_n$, which was the last one for the corresponding variable. $L_n$ is thus the last candidate of the braid in KS we were looking for.

Here again, notice that this proof works only because the existence of a link between two candidates doesn't depend on the knowledge state.

*C. Comments on the braids vs T&E theorem*

T&E is a form of blind search that is generally not accepted by advocates of pattern-based solutions (even when it allows no guessing, as in our definition of this procedure). But this theorem shows that T&E can always be replaced with a pattern based solution, more precisely with braids. The question naturally arises: can one reject T&E and nevertheless accept solutions based on braids?

As shown in section VI, resolution theories based on braids have the confluence property and many different resolution strategies can be super-imposed on them. One can decide to prefer a solution with the shorter braids available. T&E doesn't provide this (unless it is drastically modified, in ways that would make it computationally very inefficient).

Moreover, in each of these resolution theories based on braids, one can add rules corresponding to special cases, such as whips of the same lengths, and one can decide to give a natural preference to such special cases. In Sudoku, this would entail that braids which are not whips would appear in the solution of almost no random puzzle.

*D. The resolution potential of zt-braids in Sudoku*

For any CSP, the T&E vs braids theorem gives a clear theoretical answer to the question about the potential of resolution theories based on zt-braids. As the T&E procedure is very easy to implement, it also allows practical computations. We have done this for Sudoku.

We have generated 1,000,000 minimal puzzles: all of them can be solved by T&E and therefore by nrczt-braids. We already knew that nrczt-whips were enough to solve the first 10,000; checking the same thing for 1,000,000 puzzles would be too long; but it becomes easy if we consider braids instead of whips.

One should not conclude that zt-braids are a useless extension of zt-whips. We have shown that there are puzzles that cannot be solved with whips only but can be solved with braids. Said otherwise, whips are not equivalent to T&E.

VIII. CONCLUSION

Most of the general CSP solving methods [7, 8] combine a blind search algorithm with some kind of pattern-based pruning of the search graph. Here, instead of trying to solve all the instances of a CSP, as is generally the case in these methods, we have tried to push the purely pattern-based approach to its limits. In Part I, we have defined a general framework for this purpose and in Part II, we have introduced three powerful patterns, bivalue-chains, zt-whips and zt-braids. We have shown that, for any CSP, zt-braids are able to replace one level of Trial-and-Error.

We have applied this framework to the Sudoku CSP and shown that whips (resp. braids) can solve all the puzzles taken from a random sample of 10,000 (resp. 1,000,000). Nevertheless a few puzzles are known to defy both of these patterns.